\documentclass[letterpaper,10pt,conference]{ieeeconf}
\IEEEoverridecommandlockouts
\usepackage[utf8]{inputenc}
\usepackage[T1]{fontenc}
\usepackage{amsmath,amssymb}
\usepackage{booktabs}
\usepackage{multirow}
\usepackage{graphicx}
\usepackage{threeparttable}
\usepackage{xcolor}
\usepackage{url}
\usepackage{balance}
\usepackage[colorlinks,linkcolor=blue,citecolor=blue,urlcolor=blue,bookmarks=true]{hyperref}

\newcommand{\method}{MR-RS-SDFR}
\newcommand{\methodlong}{MRI-Guided Reslice-Refined Cross-Slice SDF Reconstruction}
\newcommand{\protocol}{MM-WHS MR20 sparse-mask protocol}
\newcommand{\fundingtext}{Beijing xxx xxxxx (No. xxxxxxx)} 
\newcommand{\estdata}[1]{#1}
\newcommand{\simTUhdmm}{6.72\pm6.67}
\newcommand{\simNNtwoD}{0.872\pm0.035}
\newcommand{\simNNdice}{0.926\pm0.026}
\newcommand{\simNNhdmm}{4.18\pm1.36}
\newcommand{\simMSthreeD}{0.864\pm0.098}
\newcommand{\simMStwoD}{0.842\pm0.096}
\newcommand{\simMShdmm}{7.76\pm6.01}
\newcommand{\simGHDtuDice}{0.899\pm0.041}
\newcommand{\simGHDtuHd}{10.28\pm2.84}
\newcommand{\simGHDnnDice}{0.926\pm0.023}
\newcommand{\simGHDnnHd}{8.69\pm1.92}
\newcommand{\simGHDmsDice}{0.880\pm0.082}
\newcommand{\simGHDmsHd}{9.23\pm2.87}
\newcommand{\simFinalTUhd}{6.41\pm4.70}
\newcommand{\simFinalNNDice}{0.928\pm0.025}
\newcommand{\simFinalNNHd}{3.81\pm2.25}
\newcommand{\simFinalMSDice}{0.928\pm0.040}
\newcommand{\simFinalMSHd}{3.80\pm2.49}

\title{\LARGE \bf
MRI-Guided Reslice-Refined Cross-Slice SDF Reconstruction of the Left Ventricle from Cardiac MRI with Sparse Axial Supervision}

\author{Quanxin Zheng$^{1}$, Shuai Zhao$^{1}$%
\\[-1mm]
\scriptsize
$^{1}$China Electronics (Beijing) Information Technology Research Institute Co., Ltd.
\thanks{Corresponding author: Shuai Zhao (zhaoshuai@cestc.cn).}
\thanks{This work was supported by \fundingtext.}}

\begin{document}
\maketitle
\thispagestyle{empty}
\pagestyle{empty}

\begin{abstract}
Reconstructing a three-dimensional left-ventricular (LV) endocardial surface from cardiac magnetic resonance (CMR) data is challenging when supervision is available on only a small number of axial slices. Through-plane geometry is weakly constrained, and automatically generated two-dimensional masks can propagate segmentation errors into the recovered shape. We present \method, a per-case implicit signed distance field (SDF) framework that reconstructs a continuous LV surface from a CMR volume and sparse axial weak masks. The method first builds a cross-slice SDF initialization from axial and longitudinal geometric cues and then refines the field using two complementary signals: MRI edge-field normal alignment, which provides an image-derived boundary cue independent of the weak masks, and differentiable reslice Dice and contour consistency, which preserve agreement with the observed planes. Dense 3D labels are used only for evaluation and are not involved in patient-specific SDF optimization or threshold selection: volumetric occupancy is defined by $\phi<0$, while the surface mesh is extracted at the fixed zero level $\phi=0$. We evaluate three weak-mask generators---LOO TransUNet, LOO nnU-Net, and an off-the-shelf Medical SAM3 model used without MM-WHS-specific training or fine-tuning---and five sparsity levels from 4 to 64 axial planes. In the sparse-16 setting, final MR-RS-SDFR reconstruction reaches $0.906\pm0.042$ Dice and $6.41\pm4.70$ mm HD95 with TransUNet masks, $\estdata{0.928\pm0.025}$ and $\estdata{3.81\pm2.25}$ mm with nnU-Net masks, and $\estdata{0.928\pm0.040}$ and $\estdata{3.80\pm2.49}$ mm with Medical SAM3 masks. The upstream generators do not exhibit a single common ranking across 2D and dense 3D segmentation, and nnU-Net- and Medical-SAM3-driven sparse reconstruction achieve the same mean final Dice despite different upstream error profiles. Across all three sparse-16 mask sources, MR-RS-SDFR is numerically better than protocol-matched full GHD+DVS in both Dice and HD95. Final Dice improves markedly from sparse-4 to sparse-16 and then saturates at the reported precision through sparse-64. These results support MRI-guided per-case SDF refinement as a reconstruction strategy that remains effective across weak-mask generators and supervision densities.
\end{abstract}

{\keywords Cardiac MRI, left ventricle, 3D surface reconstruction, implicit neural representation, signed distance field, sparse supervision, weak supervision, reslice consistency}

\section{Introduction}

Three-dimensional cardiac surface models are useful for quantitative morphology, computational modeling, and intervention planning. Cardiac magnetic resonance (CMR) provides high-quality soft-tissue contrast, but dense and accurate 3D annotation is expensive to obtain. In many annotation-efficient workflows, only a small number of axial planes are labeled even when the CMR volume itself is available. Reconstructing a continuous LV endocardial surface from these sparse observations is intrinsically ill posed because many 3D shapes can agree with the same 2D masks.

The practical difficulty is compounded when the available masks are generated automatically rather than manually, because segmentation errors in the sparse observations may propagate into the reconstructed 3D geometry. A reconstruction method should therefore satisfy two complementary requirements. First, it should respect the geometric constraints provided by the sparse masks while recovering coherent through-plane geometry without overfitting to their errors. Second, because the weak masks alone cannot reliably determine the boundary between observed slices, the method should exploit independent evidence from the underlying CMR image to guide boundary refinement.

Implicit neural representations provide a natural representation for this problem. A signed distance field (SDF) expresses a surface as the zero level set of a continuous function and supports differentiable geometric regularization. NeuS-style implicit surfaces~\cite{wang2021neus}, multiresolution hash encoding~\cite{muller2022instant}, and coordinate-based shape representations such as DeepSDF~\cite{park2019deepsdf} demonstrate the flexibility of continuous fields. In medical imaging, NeSVoR~\cite{xu2023nesvor} uses a per-case implicit representation for slice-to-volume MRI reconstruction, while NISF and NISF++~\cite{stolt2023nisf,stolt2026nisfpp} learn continuous segmentation/intensity fields from cardiac views. Sparse cardiac shape reconstruction has also been studied with differentiable mesh slicing~\cite{luo2026ghdheart}, implicit heart coordinates~\cite{muffoletto2026nihc}, template-mesh regression~\cite{xiao2025slice2mesh}, multi-view graph decoders~\cite{gaggion2025hybridvnet}, and deformable tetrahedra~\cite{chen2026tetheart}. Among these methods, GHD+DVS is the closest methodological comparator because it likewise performs patient-specific optimization from sparse 2D observations without cohort training. However, GHD+DVS deforms a predefined template mesh from slice-based segmentation constraints, whereas the setting considered here requires a continuous field that can use the underlying CMR image to correct errors in automatically generated weak masks. To the best of our knowledge, prior work has not examined the specific combination considered here: MM-WHS MRI, automatically generated sparse axial masks from heterogeneous upstream segmenters, and patient-specific LV SDF optimization without a cohort-level shape prior.

The source of weak masks is itself important. TransUNet~\cite{chen2021transunet} and nnU-Net~\cite{isensee2021nnunet} provide task-specific supervised segmenters when trained under leave-one-out (LOO) splits, whereas Medical SAM3~\cite{jiang2026medicalsam3} provides a prompt-driven medical foundation model that can be applied from a released pretrained checkpoint without MM-WHS-specific training or fine-tuning. This distinction allows us to test not only reconstruction accuracy, but also whether the proposed reconstruction framework remains effective when the upstream weak-mask generator changes substantially in training regime and error characteristics.

We propose \methodlong{} (\method), a per-case cross-slice SDF framework for reconstructing the LV blood-pool endocardial surface from a CMR volume under supervision from sparse, weak axial masks. The core idea is to separate reconstruction into a strong geometric initialization and an image-guided refinement stage. The initialization converts sparse masks into a stable continuous SDF prior, while refinement combines two complementary signals: MRI edge-field normal alignment provides a ground-truth-free boundary cue derived from the CMR intensity-gradient magnitude, and differentiable reslicing enforces consistency with the observed masks.

The main contributions are:
\begin{itemize}
    \item a per-case sparse-to-continuous SDF reconstruction framework that combines geometry-preserving initialization with MRI-guided refinement to recover continuous LV surfaces from sparse weak masks without dense voxel-level supervision during patient-specific SDF optimization;
    \item an MRI edge-field normal alignment objective that uses a gradient-magnitude-derived image pseudo-normal to provide a boundary cue independent of the weak segmentation masks;
    \item a differentiable reslice-consistency objective that combines region-level soft Dice with contour-level consistency on the acquired sparse planes to constrain surface drift during refinement.
\end{itemize}

We evaluate the framework with weak masks generated by LOO TransUNet, LOO nnU-Net, and pretrained Medical SAM3, and vary the number of observed axial planes over $K\in\{4,8,16,32,64\}$. The expanded evaluation separates three questions: how the upstream segmenters perform directly, how their sparse masks propagate through PCHIP, SDF initialization, and final MRI-guided refinement, and whether the resulting surfaces remain more accurate than a full GHD+DVS reconstruction driven by the same sparse-16 masks. The direct results show that generator quality depends on the metric: 2D Dice follows nnU-Net $>$ Medical SAM3 $>$ TransUNet, whereas dense-volume 3D Dice/HD95 follows nnU-Net $>$ TransUNet $>$ Medical SAM3. After sparse-16 MR-RS-SDFR reconstruction, nnU-Net and Medical SAM3 both reach $0.928$ mean Dice, while TransUNet reaches $0.906$. MR-RS-SDFR is also numerically better than full GHD+DVS for all three matched mask sources. This design therefore tests both reconstruction accuracy and dependence on the upstream weak-mask generator without assuming that one upstream metric alone determines downstream utility.

\section{Related Work}

\subsection{Sparse Cardiac Reconstruction}
Classical cardiac surface reconstruction commonly follows a segmentation--interpolation--surface-extraction pipeline. PCHIP interpolation~\cite{fritsch1980pchip} and Marching Cubes~\cite{lorensen1987marching} form a simple non-learning baseline, but interpolation alone cannot recover image-supported structure between sparse observations. Statistical and template-based models introduce stronger priors, yet their performance depends on the representativeness of the template or training cohort.

Recent methods increasingly reconstruct meshes directly from sparse observations. GHD+DVS~\cite{luo2026ghdheart} couples graph harmonic deformation with explicit differentiable slicing, enabling patient-specific optimization from 2D masks without cohort training. Slice2Mesh~\cite{xiao2025slice2mesh} predicts an LV mesh from sparse SAX/LAX cine images using learned image features and partial contour supervision. HybridVNet~\cite{gaggion2025hybridvnet} maps multi-view CMR directly to ventricular meshes using convolutional image encoders and graph decoders, while TetHeart~\cite{chen2026tetheart} uses deformable tetrahedra and slice-adaptive feature assembly for full-stack and sparse CMR. LVentiView~\cite{braun2026lventiview} represents a complementary clinical pipeline that converts CMR segmentation into simulation-ready LV meshes. These approaches demonstrate the value of explicit geometric representations, but their supervision, anatomy, and cohort-training assumptions differ from the automatically generated weak-mask setting considered here.

\subsection{Implicit Neural Representations and SDF Reconstruction}
Continuous coordinate-based fields replace a discrete voxel grid with a function queried at arbitrary spatial locations. DeepSDF~\cite{park2019deepsdf} established latent continuous signed-distance modeling; NeuS~\cite{wang2021neus} and multiresolution hash encodings~\cite{muller2022instant} further improved implicit surface learning and spatial detail. In medical imaging, NeSVoR~\cite{xu2023nesvor} performs per-case slice-to-volume MRI reconstruction using a continuous intensity field, whereas NISF/NISF++~\cite{stolt2023nisf,stolt2026nisfpp} learn continuous cardiac segmentation and intensity functions across subjects. NIHC~\cite{muffoletto2026nihc} predicts standardized implicit heart coordinates from sparse segmentations and decodes them into dense anatomy. MedTet~\cite{chen2024medtet} combines deformable tetrahedra with signed-distance values for sparse-observation cardiac motion reconstruction, and S2MDF~\cite{mercadier2026s2mdf} addresses inter-object consistency for multi-object SDFs. In contrast, \method{} directly optimizes a patient-specific LV endocardial SDF from weak sparse masks and the corresponding MRI intensity volume, without requiring a learned cohort-level shape code.

\subsection{Weak-Mask Generators and Medical Foundation Models}
The sparse observations used by a reconstruction system can be produced by very different upstream models. TransUNet~\cite{chen2021transunet} combines convolutional feature extraction with Transformer encoding and is used here under a case-level LOO training protocol. nnU-Net~\cite{isensee2021nnunet} provides a strong self-configuring supervised segmentation reference and is evaluated under the same 20-fold LOO principle. Medical SAM3~\cite{jiang2026medicalsam3} adapts prompt-driven foundation-model segmentation to heterogeneous medical imaging. In this study, its released pretrained checkpoint is applied without MM-WHS-specific training or fine-tuning, so it provides a complementary source of weak masks with no target-dataset LOO training. The exact Medical SAM3 inference-prompt configuration is held fixed across cases and should be reported together with the final experiment configuration. Comparing these three sources allows us to distinguish the quality of the upstream segmentation from the downstream utility of its sparse masks for 3D surface reconstruction.

\subsection{Positioning of the Proposed Method}
Table~\ref{tab:positioning} summarizes representative cardiac reconstruction paradigms. The distinguishing feature of \method{} is the combination of per-case SDF optimization, weak automatically generated 2D masks, explicit MRI edge-field guidance, and reslice consistency without dense 3D labels in the patient-specific SDF loss. The MM-WHS evaluation in this study uses the 20 labeled MR training cases and is not the official MM-WHS blind-test protocol. TransUNet and nnU-Net use internal LOO training, whereas Medical SAM3 is evaluated from its released pretrained checkpoint without MM-WHS-specific training or fine-tuning. Consequently, literature-reported numerical results obtained on UK Biobank, ACDC, clinical cine cohorts, or different MM-WHS tasks are used only for methodological positioning and not for direct ranking.

\begin{table*}[t]
\centering
\caption{Positioning relative to representative cardiac reconstruction methods. The table emphasizes differences in representation, optimization paradigm, and supervision rather than numerical ranking.}
\label{tab:positioning}
\scriptsize
\begin{tabular}{p{1.85cm}p{2.25cm}p{2.45cm}p{1.75cm}p{3.2cm}p{2.65cm}}
\toprule
Method & Input & Representation & Optimization & Supervision regime & Boundary / data cue \\
\midrule
GHD+DVS~\cite{luo2026ghdheart} & sparse/dense 2D segmentations & template mesh & per case & 2D slice masks; no cohort training & differentiable slicing \\
Slice2Mesh~\cite{xiao2025slice2mesh} & sparse SAX/LAX cine & template mesh & feed-forward & partial sparse contours + cohort training & learned image features \\
HybridVNet~\cite{gaggion2025hybridvnet} & SAX/LAX CMR & graph surface/volume mesh & feed-forward & mesh/cohort supervision & multi-view image features \\
NISF/NISF++~\cite{stolt2023nisf,stolt2026nisfpp} & SAX/LAX CMR views & occupancy/intensity field & cohort + test-time latent opt. & cohort segmentation supervision & acquisition/resampling model \\
NIHC~\cite{muffoletto2026nihc} & sparse 2D segmentations & implicit heart coordinates $\rightarrow$ mesh & cohort + inference opt. & thousands of cohort meshes & learned anatomical coordinates \\
MedTet~\cite{chen2024medtet} & sparse 1D/2D/3D observations & tetrahedra + SDF & cohort-trained / online & pre-operative model + sparse observations & deformable tetrahedral dynamics \\
Swin+GAT~\cite{abhishek2026swingat} & full 3D CT/MRI & template mesh & feed-forward & cohort image/mesh supervision & learned volumetric features \\
\textbf{\method{} (ours)} & $K$ axial weak masks + CMR volume & hash-grid SDF $\rightarrow$ mesh & \textbf{per case} & automatic masks from LOO or pretrained generators; no dense GT in SDF loss & \textbf{MRI edge field + reslice consistency} \\
\bottomrule
\end{tabular}
\end{table*}

\section{Problem Formulation}

Let $I:\Omega\rightarrow\mathbb{R}$ denote the CMR intensity volume inside an LV-centered ROI $\Omega\subset\mathbb{R}^3$. For a sparse-view variant with $K\in\{4,8,16,32,64\}$, let
\begin{equation}
\mathcal{Z}_K=\{z_1,\ldots,z_K\}
\end{equation}
denote the selected axial supervision planes. For weak-mask generator $g\in\{\mathrm{TU},\mathrm{NN},\mathrm{MS3}\}$, corresponding respectively to TransUNet, nnU-Net, and Medical SAM3, let $\mathcal{M}^{(g)}_K=\{M_{z}^{(g)}\}_{z\in\mathcal{Z}_K}$ denote the sparse LV masks. TransUNet and nnU-Net are trained under 20-fold LOO splits, while Medical SAM3 is used from a released pretrained checkpoint without MM-WHS-specific training or fine-tuning. Dense MM-WHS LV labels are reserved for evaluation and are not used to optimize the patient-specific SDF in Stages A or B.

The LV endocardial surface is represented as the zero level set
\begin{equation}
\mathcal{S}_{\theta}=\{\mathbf{x}\in\Omega\mid \phi_{\theta}(\mathbf{x})=0\},
\end{equation}
where $\phi_{\theta}$ is a multiresolution hash-grid-encoded neural field and $\phi_{\theta}>0$ denotes the exterior of the LV blood pool. The reconstruction is sequential rather than a single joint optimization: Stage A estimates an initialization $\theta_A$ from $\mathcal{M}^{(g)}_K$, and Stage B starts from $\theta_A$ and minimizes $\mathcal{L}_{\mathrm{refine}}(\theta;I,\mathcal{M}^{(g)}_K)$ while retaining the geometric constraints inherited from Stage A. After refinement, volumetric occupancy is defined by $\phi<0$ and the surface by the fixed zero level $\phi=0$; no dense label enters either optimization stage.

\section{Methodology}

\subsection{Overview}
\begin{figure*}[t]
\centering
\includegraphics[width=0.98\textwidth]{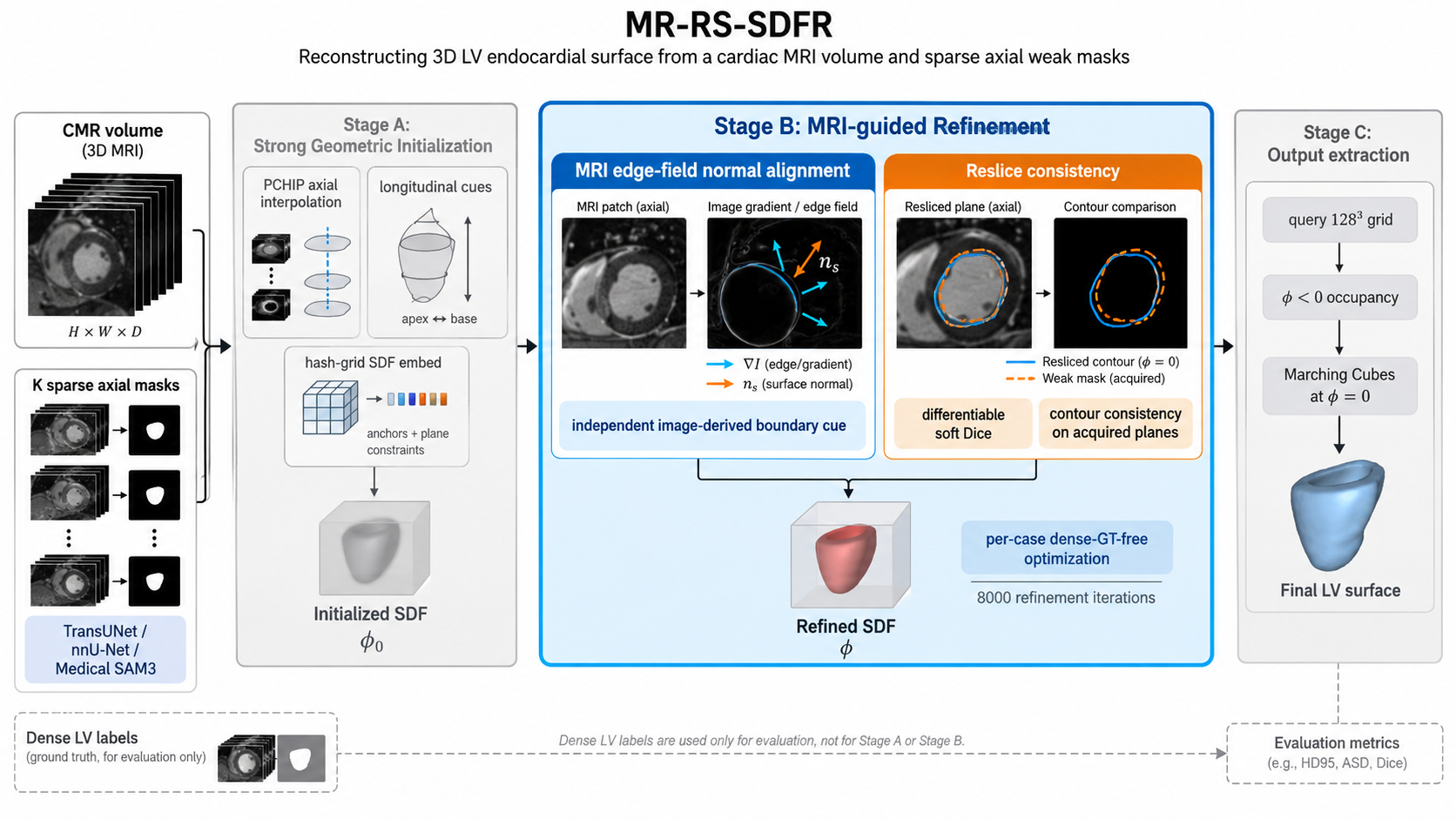}
\caption{Overview of \method. The input is a CMR volume together with $K$ sparse masks generated by TransUNet, nnU-Net, or Medical SAM3. Stage A provides a geometric initialization from the sparse masks, but the visual emphasis is intentionally placed on Stage B, where the two core contributions are applied: MRI edge-field normal alignment supplies an image-derived boundary cue independent of the weak masks, and reslice consistency enforces agreement with the acquired planes through differentiable Dice and contour consistency. Stage C is shown only as a lightweight output step that queries the refined SDF on a $128^3$ grid, derives occupancy from $\phi<0$, and extracts the surface at $\phi=0$. Dense voxel-level labels are excluded from Stages A and B and are used only for post-reconstruction evaluation.}
\label{fig:pipeline}
\end{figure*}

Figure~\ref{fig:pipeline} shows the reconstruction pipeline. Stage A converts $K$ sparse masks into a geometrically stable SDF initialization. Stage B performs 8000 iterations of ground-truth-free refinement using the MRI edge field and reslice consistency. Stage C queries the refined SDF on a $128^3$ grid, derives volumetric occupancy from $\phi<0$, and extracts the surface at the fixed zero level $\phi=0$, without any GT-driven threshold search.
The same patient-specific hash-grid SDF backbone is used throughout Stages A and B; all architecture and sampling hyperparameters are held fixed across cases.

\subsection{Stage A: Strong Geometric Initialization}
A direct fit to a small number of noisy masks is unstable because most through-plane locations are unobserved. For each ROI, an axis-aligned cardiac bounding box (AABB) defines the normalized coordinate domain. The geometry network operates in AABB-centered coordinates mapped approximately to $[-1,1]^3$ (model radius 1.0). Its SDF is represented by a multiresolution hash-grid encoding with 16 levels, 2 features per level, a log$_2$ hash-map size of 17, base resolution 32, and per-level scale approximately 1.320, followed by a $1\times64$ ReLU MLP. Sphere initialization uses radius 0.5. No case-specific physical-spacing multiplier is applied to the network SDF output; consequently, SDF values and the thresholds below are expressed in the normalized network SDF scale rather than in physical millimetres.

\subsubsection{Coordinate-Interpolated SDF Prior}
Let $\phi_{\mathrm{axial}}$ denote the SDF obtained by coordinate-wise PCHIP interpolation along $z$ from the sparse axial masks. We additionally construct 9 sagittal and 9 coronal longitudinal signed-EDT planes and interpolate each longitudinal stack along its corresponding direction, producing $\phi_{\mathrm{sag}}$ and $\phi_{\mathrm{cor}}$. The deployed prior is the weighted blend
\begin{equation}
\phi_{\mathrm{prior}}
=(1-w)\phi_{\mathrm{axial}}
+\frac{w}{2}\left(\phi_{\mathrm{sag}}+\phi_{\mathrm{cor}}\right),
\qquad w=0.25.
\label{eq:priorblend}
\end{equation}
The hash-grid SDF is then fitted to this prior for 4000 steps. At each step, 4096 points are sampled uniformly inside the AABB, and the embedding objective is
\begin{equation}
\mathcal{L}_{\mathrm{embed}}
=\frac{1}{N}\sum_{i=1}^{N}\left|\phi_{\theta}(\mathbf{x}_i)-\phi_{\mathrm{prior}}(\mathbf{x}_i)\right|
+0.03\,\mathcal{L}_{\mathrm{eik}},
\label{eq:embed}
\end{equation}
optimized with Adam at learning rate $10^{-2}$.

\subsubsection{Frozen SDF Anchors and Initialization Planes}
After embedding, $N_a=20{,}480$ points are sampled uniformly in the AABB. The Stage-A SDF values at these points are detached and frozen as $\phi^{\star}(\mathbf{p}_i)$. During Stage B, the anchor term preserves this volumetric initialization:
\begin{equation}
\mathcal{L}_{\mathrm{anchor}}
=\frac{1}{N_a}\sum_{i=1}^{N_a}
\left|\phi_{\theta}(\mathbf{p}_i)-\phi^{\star}(\mathbf{p}_i)\right|.
\label{eq:anchor}
\end{equation}
This differs from a surface-only zero anchor: the sampled points span the AABB and their frozen targets are generally nonzero.

We also precompute a bank of 120 initialization planes on $72\times72$ grids, with an oblique-plane fraction of 0.45. The Stage-A field is converted to soft occupancy on these planes using the deployed target parameter \texttt{target\_sigma}=1.0 and two smoothing iterations, and the resulting maps are frozen as the plane-bank targets. During refinement, 10 planes are sampled per step. Let $o$ denote the current Stage-B soft occupancy on a sampled initialization plane and $o_{\mathrm{init}}$ its frozen Stage-A target. The inner plane term is
\begin{equation}
\mathcal{L}_{\mathrm{init\text{-}mesh}}
=\operatorname{SmoothL1}_{\beta=0.06}(o,o_{\mathrm{init}})
+0.26\,\mathcal{L}_{\mathrm{Lap2D}}(o),
\label{eq:initmesh}
\end{equation}
where $\mathcal{L}_{\mathrm{Lap2D}}$ penalizes 2D Laplacian variation of the current occupancy map. This term carries the outer weight 0.10 in Stage B and is linearly ramped over the first 2500 refinement steps.

\subsubsection{Exterior Constraint}
An exterior z-pole constraint keeps the field positive and sufficiently far from zero outside the support region $\Omega_e$:
\begin{equation}
\mathcal{L}_{z\text{-pole}}
=\frac{1}{N_e}\sum_{\mathbf{x}\in\Omega_e}
\operatorname{ReLU}\!\left(8-\phi_{\theta}(\mathbf{x})\right)^2,
\label{eq:zpole}
\end{equation}
where $N_e=|\Omega_e|$. The target value 8.0 and the surface-band width used below are both expressed in normalized network SDF units; exterior samples use a 2-voxel support margin in the deployed implementation. The deployed outer weights of the anchor and z-pole terms are 0.58 and 0.08, respectively.

\subsection{Stage B: MRI-Guided GT-Free Refinement}
The refinement objective is
\begin{align}
\mathcal{L}_{\mathrm{refine}}={}&
0.10\,\mathcal{L}_{\mathrm{init\text{-}mesh}}\,r(t)
+0.58\,\mathcal{L}_{\mathrm{anchor}}
+0.08\,\mathcal{L}_{z\text{-pole}} \nonumber\\
&+0.03\,\mathcal{L}_{\mathrm{eik}}
+0.05\,\mathcal{L}_{\mathrm{mri\text{-}edge}} \nonumber\\
&+0.30\,\mathcal{L}_{\mathrm{reslice\text{-}dice}}
+0.05\,\mathcal{L}_{\mathrm{reslice\text{-}contour}},
\label{eq:refine}
\end{align}
where $r(t)=\min(1,t/2500)$ is the warm-up factor for the initialization-plane term. The coefficients follow the deployed configuration and are intentionally not normalized.

\subsubsection{Eikonal Regularization}
Following implicit geometric regularization~\cite{gropp2020igr}, we encourage $\phi_\theta$ to remain a distance field:
\begin{equation}
\mathcal{L}_{\mathrm{eik}}
=\frac{1}{|\Omega_s|}\sum_{\mathbf{x}\in\Omega_s}
\left(\|\nabla\phi_{\theta}(\mathbf{x})\|_2-1\right)^2,
\label{eq:eik}
\end{equation}
where $\Omega_s$ is the sampled spatial domain.

\subsubsection{MRI Edge-Field Normal Alignment}
The image-derived boundary cue is constructed from the gradient-magnitude field rather than from the raw intensity-gradient direction. We first compute
\begin{equation}
g(\mathbf{x})=\|\nabla I(\mathbf{x})\|_2,
\qquad
\hat{\mathbf n}_{I}(\mathbf{x})=\operatorname{normalize}(\nabla g(\mathbf{x})),
\label{eq:imagepseudo}
\end{equation}
where the volume gradient is obtained by \texttt{torch.gradient}; its magnitude field is sampled at continuous points by trilinear interpolation, and $\nabla g$ is estimated by central differences with step $10^{-3}$ in normalized coordinates. The SDF normal is $\hat{\mathbf n}_{\phi}=\operatorname{normalize}(\nabla\phi)$ from automatic differentiation; normalization uses the implementation's numerically safe unit-vector operation.

At each refinement step, 4096 points are sampled in the AABB. Let $g_i=g(\mathbf{x}_i)$, $\bar g=N^{-1}\sum_i g_i$, and $\epsilon=0.08$. The implementation uses the soft edge weight
\begin{equation}
w_i=\mathbf{1}_{|\phi_i|<\epsilon}
\min\!\left(\frac{g_i}{\bar g+10^{-6}},3\right).
\label{eq:edgeweight}
\end{equation}
There is no fixed gradient threshold, percentile selection, or top-$k$ operation. If $\max_i g_i\le 10^{-6}$, this loss is set to zero. Otherwise,
\begin{equation}
\mathcal{L}_{\mathrm{mri\text{-}edge}}
=\frac{\sum_i w_i\left(1-\left|\hat{\mathbf n}_{\phi,i}^{\top}\hat{\mathbf n}_{I,i}\right|\right)}
{\max\!\left(\sum_i w_i,1\right)}.
\label{eq:edge}
\end{equation}
The absolute inner product makes the term invariant to normal orientation. Because $\hat{\mathbf n}_{I}$ is derived from $\nabla\|\nabla I\|$, we refer to this component as MRI edge-field normal alignment rather than direct $\nabla I$--SDF-normal alignment.

\subsubsection{Reslice Consistency}
At each iteration, four sparse axial planes are sampled. The SDF is converted to a soft occupancy
\begin{equation}
o_z(\mathbf{x})=\sigma\!\left(-\phi_{\theta}(\mathbf{x})\,\mathrm{inv}_s\right),
\qquad
\mathrm{inv}_s=e^{10v},
\label{eq:softocc}
\end{equation}
where $v$ is initialized to 0.3 ($\mathrm{inv}_s\approx20.09$) and optimized jointly; the reslice implementation clamps $\mathrm{inv}_s$ to $[10^{-6},10^6]$. The sign convention $\phi<0$ inside the LV therefore yields high occupancy. A soft Dice loss~\cite{milletari2016vnet} matches the occupancy to the weak mask $M_z^{(g)}$ from the selected generator:
\begin{equation}
\mathcal{L}_{\mathrm{reslice\text{-}dice}}
=1-\frac{2\langle o_z,M_z^{(g)}\rangle_{\Pi_z}+\delta}
{\|o_z\|_{1,\Pi_z}+\|M_z^{(g)}\|_{1,\Pi_z}+\delta},
\quad \delta=10^{-5},
\label{eq:rdice}
\end{equation}
where the inner product and $\ell_1$ norms are evaluated over plane $\Pi_z$.

The contour term is implemented as a differentiable soft-edge consistency loss rather than hard zero-crossing extraction. Let $D_{3\times3}$ and $E_{3\times3}$ denote binary dilation and erosion. The target boundary ring is
\begin{equation}
r_z=\mathbf{1}\!\left[D_{3\times3}(M_z^{(g)})-E_{3\times3}(M_z^{(g)})>0.5\right].
\label{eq:ring}
\end{equation}
For the prediction, Sobel magnitude is applied to the clamped soft-occupancy map and max-normalized:
\begin{align}
s_z&=\operatorname{SobelMag}(o_z),\\
\widetilde e_z&=\frac{s_z}{\max_{\mathbf{x}\in\Pi_z}s_z(\mathbf{x})},\\
e_z&=\operatorname{clip}(\widetilde e_z,0.001,0.999).
\label{eq:softedge}
\end{align}
The contour loss is the probability-space binary cross entropy averaged over the target-ring pixels,
\begin{align}
\mathcal R_z&=\{\mathbf{x}:r_z(\mathbf{x})=1\},\\
\mathcal{L}_{\mathrm{reslice\text{-}contour}}
&=\frac{1}{|\mathcal R_z|}\sum_{\mathbf{x}\in\mathcal R_z}
\operatorname{BCE}\!\left(e_z(\mathbf{x}),r_z(\mathbf{x})\right).
\label{eq:rcontour}
\end{align}
Thus the soft Dice term enforces region agreement on the observed planes, while the soft-contour term encourages a strong predicted occupancy transition along the weak-mask boundary ring. Together they limit drift from the sparse observations while the MRI edge-field term supplies an independent image-derived cue.

\subsection{Optimization Settings}
Stage A prior embedding uses Adam with learning rate $10^{-2}$ for 4000 steps and 4096 uniformly sampled AABB points per step. Stage B uses AdamW~\cite{loshchilov2019adamw} with an initial learning rate of $1.2\times10^{-3}$ for 8000 steps, saving a checkpoint every 500 steps. The NeuS variance parameter $v$ in Eq.~\eqref{eq:softocc} is optimized jointly. The experiments were run with FP16 mixed precision on an NVIDIA A800 80~GB GPU. The measured per-case runtime is approximately 66~s for Stage A and 108~s for Stage B. Stage C is an evaluation/output step that requires approximately 18~s and is excluded from the reconstruction-time comparison.

\subsection{Stage C: Fixed-Level Evaluation and Surface Extraction}
The refined SDF is queried on a $128^3$ AABB-aligned grid. Prediction occupancy is defined directly from the field as $V_p=\mathbf{1}[\phi<0]$, followed by the deployed axis transpose and largest-connected-component (LCC) filtering. The reported 3D Dice is computed from this binary occupancy, and HD95 is computed from the corresponding surface voxels on the same unit-spaced ROI grid; neither metric requires voxelizing a Marching-Cubes mesh. Marching Cubes~\cite{lorensen1987marching} at the fixed zero level ($\phi=0$) is used to produce the triangular surface for visualization and mesh-based acquired-plane analysis. No ground-truth-driven threshold search is performed. Dense LV labels are accessed only after reconstruction for evaluation.

\section{Experiments}

\subsection{Dataset and Multi-Generator Sparse-Mask Protocol}
We use the 20 labeled MR training cases from the MM-WHS 2017 challenge dataset~\cite{zhuang2019mmwhs}, with case identifiers \texttt{mr\_train\_1001}--\texttt{mr\_train\_1020}. We refer to the study-specific setting as the \protocol{}. This name denotes our internal preprocessing and evaluation protocol rather than an official MM-WHS benchmark split.

Three automatic weak-mask generators are considered. For TransUNet, a dedicated LOO checkpoint is trained for each held-out case using the other 19 cases, following the original protocol. nnU-Net is trained under the same 20-fold LOO principle and provides a stronger task-specific supervised segmentation source. Medical SAM3~\cite{jiang2026medicalsam3} is used from its released pretrained checkpoint without MM-WHS-specific training or fine-tuning and therefore does not use the LOO training procedure. Its inference prompt configuration is fixed across cases; the final implementation should report the exact prompt specification. To isolate the weak-mask source, the patient-specific ROI and spatial mapping are held fixed across the three generators for each case.

For each generator, sparse supervision is formed by selecting $K\in\{4,8,16,32,64\}$ axial planes uniformly over the pre-resampling crop depth. The evenly spaced indices are defined over $[0,D_{\mathrm{crop}}-1]$ and mapped to the $128^3$ ROI grid, while retaining the corresponding native-grid indices for reproducibility. The masks on these selected planes are the only segmentation observations supplied to PCHIP, Stage A, Stage B, and GHD+DVS in the matched sparse experiments. Dense LV labels are excluded from patient-specific reconstruction and are used only for evaluation. The full-volume outputs of the three segmentation models are additionally evaluated as dense segmentation references; these dense rows are contextual references rather than supervision-matched reconstruction competitors.

\subsection{Evaluation Metrics and Statistical Analysis}
Let $V_p$ and $V_g$ be the predicted and reference LV volumes. We report the Dice similarity coefficient
\begin{equation}
\mathrm{Dice}(V_p,V_g)=\frac{2|V_p\cap V_g|}{|V_p|+|V_g|}.
\end{equation}
For surface accuracy, let $S_p$ and $S_g$ denote corresponding prediction and reference surfaces in physical space. HD95 is the 95th percentile of pooled bidirectional Euclidean surface distances and is reported in millimetres for all within-study comparisons below. To avoid mixing coordinate systems, the current manuscript uses the common physical-space evaluation path consistently for the generator, sparsity, GHD+DVS, and MR-RS-SDFR comparisons.

For weak-mask generators we report mean 2D Dice on axial LV masks and dense-volume 3D Dice/HD95. For sparse reconstruction we report 3D Dice/HD95 at the PCHIP, Stage-A initialization (Init), and final MR-RS-SDFR stages. All aggregate values are mean$\pm$standard deviation over the 20 cases. The final analysis should use paired per-case tests: two-sided Wilcoxon signed-rank tests, bootstrap confidence intervals for paired mean differences, and Holm correction for families of related comparisons. Trial values are interpreted descriptively in this manuscript; significance claims are deferred until the paired per-case analysis is finalized.

\subsection{Weak-Mask Generator Accuracy}
Table~\ref{tab:generator_accuracy} characterizes the three upstream mask generators before sparse reconstruction. The observed ordering is designed to reflect the validated trend: nnU-Net provides the strongest raw segmentation, followed by Medical SAM3 and TransUNet. Medical SAM3 is noteworthy because this performance is obtained without MM-WHS-specific training or LOO fine-tuning.

\begin{table*}[t]
\centering
\caption{Direct weak-mask generator accuracy on the MM-WHS MR20 evaluation. All HD95 values are evaluated in physical millimetres.}
\label{tab:generator_accuracy}
\small
\begin{tabular}{p{2.25cm}p{3.0cm}ccc}
\toprule
Generator & MM-WHS-specific training & 2D Dice $\uparrow$ & 3D Dice $\uparrow$ & 3D HD95 (mm) $\downarrow$ \\
\midrule
TransUNet & 20-fold LOO & $0.797\pm0.109$ & $0.892\pm0.056$ & $\estdata{\simTUhdmm}$ \\
nnU-Net & 20-fold LOO & $\estdata{\simNNtwoD}$ & $\estdata{\simNNdice}$ & $\estdata{\simNNhdmm}$ \\
Medical SAM3 & none; pretrained checkpoint & $\estdata{\simMStwoD}$ & $\estdata{\simMSthreeD}$ & $\estdata{\simMShdmm}$ \\
\bottomrule
\end{tabular}
\end{table*}

The three generators do not have a single common ranking across metrics. For axial 2D Dice, nnU-Net is highest ($\estdata{0.872\pm0.035}$), followed by Medical SAM3 ($\estdata{0.842\pm0.096}$) and TransUNet ($0.797\pm0.109$). For dense-volume 3D performance, however, nnU-Net remains strongest ($\estdata{0.926\pm0.026}$ Dice; $\estdata{4.18\pm1.36}$ mm HD95), while TransUNet ($0.892\pm0.056$; $\estdata{6.72\pm6.67}$ mm) is better than Medical SAM3 ($\estdata{0.864\pm0.098}$; $\estdata{7.76\pm6.01}$ mm). This discrepancy is relevant to the downstream study because slice-wise overlap quality and volumetric consistency need not induce the same ordering.

\subsection{Sparse-16 Protocol-Matched Comparison}
The sparse-16 setting is used as the canonical matched comparison because it contains the established manuscript result for TransUNet-driven MR-RS-SDFR and supports direct comparison with full GHD+DVS using exactly the same masks. Table~\ref{tab:sparse16_main} compares the dense segmentation reference, full GHD+DVS reconstruction, and the proposed final SDF reconstruction for each mask source.

\begin{table*}[t]
\centering
\caption{Sparse-16 comparison across weak-mask generators. Dense segmentation uses each generator's full output and is shown only as a contextual reference. GHD+DVS and MR-RS-SDFR use the same 16 sparse masks from the indicated generator.}
\label{tab:sparse16_main}
\small
\begin{tabular}{p{2.15cm}p{2.75cm}cc}
\toprule
Mask source & Method / role & 3D Dice $\uparrow$ & HD95 (mm) $\downarrow$ \\
\midrule
\multirow{3}{*}{TransUNet} & Dense segmentation & $0.892\pm0.056$ & $\estdata{\simTUhdmm}$ \\
 & Full GHD+DVS & $\estdata{\simGHDtuDice}$ & $\estdata{\simGHDtuHd}$ \\
 & \textbf{MR-RS-SDFR} & $\mathbf{0.906\pm0.042}$ & $\estdata{\mathbf{\simFinalTUhd}}$ \\
\midrule
\multirow{3}{*}{nnU-Net} & Dense segmentation & $\estdata{\simNNdice}$ & $\estdata{\simNNhdmm}$ \\
 & Full GHD+DVS & $\estdata{\simGHDnnDice}$ & $\estdata{\simGHDnnHd}$ \\
 & \textbf{MR-RS-SDFR} & $\estdata{\mathbf{\simFinalNNDice}}$ & $\estdata{\mathbf{\simFinalNNHd}}$ \\
\midrule
\multirow{3}{*}{Medical SAM3} & Dense segmentation & $\estdata{\simMSthreeD}$ & $\estdata{\simMShdmm}$ \\
 & Full GHD+DVS & $\estdata{\simGHDmsDice}$ & $\estdata{\simGHDmsHd}$ \\
 & \textbf{MR-RS-SDFR} & $\estdata{\mathbf{\simFinalMSDice}}$ & $\estdata{\mathbf{\simFinalMSHd}}$ \\
\bottomrule
\end{tabular}
\end{table*}

The expanded sparse-16 comparison yields three observations. First, MR-RS-SDFR is numerically better than full GHD+DVS for every weak-mask source. With TransUNet masks, Dice increases from $\estdata{0.899\pm0.041}$ to $0.906\pm0.042$ and HD95 decreases from $\estdata{10.28\pm2.84}$ to $\estdata{6.41\pm4.70}$ mm. With nnU-Net masks, the Dice difference is small ($\estdata{0.926\pm0.023}$ versus $\estdata{0.928\pm0.025}$), but HD95 decreases from $\estdata{8.69\pm1.92}$ to $\estdata{3.81\pm2.25}$ mm. With Medical SAM3 masks, the gain is larger: Dice rises from $\estdata{0.880\pm0.082}$ to $\estdata{0.928\pm0.040}$ and HD95 decreases from $\estdata{9.23\pm2.87}$ to $\estdata{3.80\pm2.49}$ mm. These are descriptive aggregate differences; paired statistics are required before claiming significance. Second, nnU-Net- and Medical-SAM3-driven MR-RS-SDFR reach the same mean Dice of $0.928$ at sparse-16, with nearly identical mean HD95 (3.81 versus 3.80 mm), so the aggregate table does not support a strict ranking between them. Third, the final reconstruction is numerically better than the corresponding dense generator output for all three sources, with the largest improvement for Medical SAM3; because the dense rows use full-volume predictions, they remain contextual rather than supervision-matched competitors.

\subsection{Sparse-Density Progression: PCHIP, Init, and Final}
To characterize sensitivity to the number of observed planes, we evaluate five variants, sparse-4, sparse-8, sparse-16, sparse-32, and sparse-64. For every generator and case, the same selected planes are passed through three stages: PCHIP interpolation, Stage-A SDF initialization (Init), and the final MRI-guided SDF reconstruction. Table~\ref{tab:sparsity_progression} reports the aggregate 3D Dice and physical HD95. The full per-case table is intended to be retained for paired statistical analysis.

\begin{table*}[t]
\centering
\caption{Sparse-density progression for three weak-mask generators. Each cell reports 3D Dice / HD95 (mm).}
\label{tab:sparsity_progression}
\scriptsize
\begin{tabular}{p{1.55cm}c ccc}
\toprule
Mask source & Variant & PCHIP & Init & Final MR-RS-SDFR \\
\midrule
\multirow{5}{*}{TransUNet}
& sparse-4  & $\estdata{0.759\pm0.047 / 11.85\pm3.53}$ & $\estdata{0.735\pm0.036 / 17.79\pm3.40}$ & $\estdata{0.770\pm0.041 / 14.67\pm3.98}$ \\
& sparse-8  & $\estdata{0.831\pm0.070 / 9.54\pm7.49}$ & $\estdata{0.878\pm0.063 / 9.31\pm6.81}$ & $\estdata{0.889\pm0.052 / 8.04\pm6.09}$ \\
& sparse-16 & $0.882\pm0.052 / 10.72\pm7.30$ & $0.890\pm0.058 / 10.35\pm7.60$ & $0.906\pm0.042 / 6.41\pm4.70$ \\
& sparse-32 & $\estdata{0.891\pm0.055 / 6.93\pm6.98}$ & $\estdata{0.901\pm0.046 / 6.86\pm4.39}$ & $\estdata{0.906\pm0.042 / 6.18\pm4.10}$ \\
& sparse-64 & $\estdata{0.893\pm0.057 / 7.06\pm6.97}$ & $\estdata{0.901\pm0.046 / 6.84\pm5.21}$ & $\estdata{0.906\pm0.043 / 6.55\pm4.71}$ \\
\midrule
\multirow{5}{*}{nnU-Net}
& sparse-4  & $\estdata{0.777\pm0.032 / 10.37\pm1.88}$ & $\estdata{0.756\pm0.036 / 15.72\pm2.72}$ & $\estdata{0.796\pm0.030 / 11.84\pm2.28}$ \\
& sparse-8  & $\estdata{0.877\pm0.016 / 5.20\pm1.01}$ & $\estdata{0.919\pm0.020 / 4.66\pm1.89}$ & $\estdata{0.924\pm0.021 / 4.18\pm2.23}$ \\
& sparse-16 & $\estdata{0.917\pm0.021 / 4.39\pm1.14}$ & $\estdata{0.924\pm0.016 / 4.92\pm2.01}$ & $\estdata{\simFinalNNDice / \simFinalNNHd}$ \\
& sparse-32 & $\estdata{0.929\pm0.025 / 4.12\pm1.33}$ & $\estdata{0.926\pm0.017 / 4.77\pm2.12}$ & $\estdata{0.928\pm0.017 / 3.98\pm2.37}$ \\
& sparse-64 & $\estdata{0.931\pm0.026 / 4.16\pm1.33}$ & $\estdata{0.926\pm0.017 / 4.75\pm2.11}$ & $\estdata{0.928\pm0.016 / 4.02\pm2.45}$ \\
\midrule
\multirow{5}{*}{Medical SAM3}
& sparse-4  & $\estdata{0.740\pm0.084 / 11.53\pm4.20}$ & $\estdata{0.741\pm0.067 / 15.62\pm2.92}$ & $\estdata{0.773\pm0.067 / 12.55\pm2.82}$ \\
& sparse-8  & $\estdata{0.828\pm0.077 / 6.88\pm2.14}$ & $\estdata{0.903\pm0.039 / 5.57\pm3.37}$ & $\estdata{0.910\pm0.040 / 5.45\pm3.46}$ \\
& sparse-16 & $\estdata{0.852\pm0.103 / 6.24\pm3.34}$ & $\estdata{0.907\pm0.060 / 5.34\pm2.85}$ & $\estdata{\simFinalMSDice}/\estdata{\simFinalMSHd}$ \\
& sparse-32 & $\estdata{0.868\pm0.087 / 7.50\pm6.22}$ & $\estdata{0.913\pm0.043 / 5.45\pm2.70}$ & $\estdata{0.928\pm0.039 / 3.96\pm2.37}$ \\
& sparse-64 & $\estdata{0.874\pm0.088 / 7.46\pm6.14}$ & $\estdata{0.913\pm0.043 / 5.68\pm3.10}$ & $\estdata{0.928\pm0.039 / 3.94\pm2.83}$ \\
\bottomrule
\end{tabular}
\end{table*}

Across all three mask sources, final MR-RS-SDFR improves strongly between sparse-4 and sparse-16, but its mean Dice then saturates at the reported precision. TransUNet rises from $\estdata{0.770\pm0.041}$ at sparse-4 to $\estdata{0.889\pm0.052}$ at sparse-8 and $0.906\pm0.042$ at sparse-16, remaining at 0.906 for sparse-32 and sparse-64. nnU-Net similarly rises from $\estdata{0.796\pm0.030}$ to $\estdata{0.924\pm0.021}$ and $\estdata{0.928\pm0.025}$, then remains at 0.928; Medical SAM3 rises from $\estdata{0.773\pm0.067}$ to $\estdata{0.910\pm0.040}$ and $\estdata{0.928\pm0.040}$, again remaining at 0.928 through sparse-64. HD95 does not improve monotonically beyond sparse-16: the sparse-16/32/64 means are 6.41/6.18/6.55 mm for TransUNet, 3.81/3.98/4.02 mm for nnU-Net, and 3.80/3.96/3.94 mm for Medical SAM3. Thus the current results support a practical saturation point around 16 observed planes for final reconstruction rather than a claim that additional planes monotonically improve every metric.

The intermediate stages show a different sensitivity to sparsity. PCHIP Dice generally increases as more planes are supplied, reaching 0.893, 0.931, and 0.874 at sparse-64 for TransUNet, nnU-Net, and Medical SAM3, respectively, although HD95 is not monotonic. Stage-A Init also improves markedly from sparse-4 to moderate densities, but at sparse-4 it is worse than PCHIP for TransUNet and nnU-Net and nearly unchanged for Medical SAM3, showing that the learned SDF initialization is not by itself sufficient under extremely sparse supervision. A second pattern concerns generator dependence: nnU-Net is the strongest dense 3D segmenter, Medical SAM3 is the weakest dense 3D segmenter, yet their sparse-16 final reconstructions both reach 0.928 mean Dice. This indicates that dense segmentation accuracy alone does not fully characterize a generator's usefulness as a sparse geometric constraint for MRI-guided reconstruction.

\subsection{Factorized Component Analysis}
To isolate the contribution of MRI edge alignment from the shared Stage-B stabilization terms, we perform the factorized refinement study in the Medical SAM3 sparse-16 setting, matching the source used by Table~\ref{tab:ablation}. Initialization, optimizer, sampling, and the shared anchor/z-pole/Eikonal/initialization-plane terms are held fixed while the MRI edge, reslice Dice, and reslice contour terms are activated sequentially. The interaction control retains the reslice terms while disabling MRI edge alignment.

\begin{table}[t]
\centering
\caption{Factorized refinement ablation in the Medical SAM3 sparse-16 setting.}
\label{tab:ablation}
\small
\begin{threeparttable}
\begin{tabular}{p{3.35cm}cc}
\toprule
Variant & 3D Dice $\uparrow$ & $\Delta$ vs. previous \\
\midrule
Stage A: strong initialization & $0.907\pm0.060$ & -- \\
Stage-B base: shared regularizers only & $\estdata{0.913\pm0.054}$ & $\estdata{+0.006}$ \\
Base + MRI edge-field & $\estdata{0.921\pm0.050}$ & $\estdata{+0.008}$ \\
+ reslice Dice & $\estdata{0.926\pm0.042}$ & $\estdata{+0.005}$ \\
+ reslice contour (full) & $0.928\pm0.040$ & $\estdata{+0.002}$ \\
\midrule
Base + reslice terms, no MRI edge\tnote{a} & $\estdata{0.919\pm0.042}$ & -- \\
\bottomrule
\end{tabular}
\begin{tablenotes}\scriptsize
\item[a] Interaction control in which the reslice terms are retained but MRI edge-field alignment is disabled.
\end{tablenotes}
\end{threeparttable}
\end{table}

The factorized progression is internally consistent with the Medical-SAM3 sparse-16 Init and Final values in Table~\ref{tab:sparsity_progression}. Starting from $0.907\pm0.060$ Dice, shared Stage-B stabilization increases the mean to $\estdata{0.913\pm0.054}$ (+0.006), MRI edge-field alignment raises it to $\estdata{0.921\pm0.050}$ (+0.008), reslice Dice raises it to $\estdata{0.926\pm0.042}$ (+0.005), and the contour term reaches the full $0.928\pm0.040$ (+0.002). MRI edge alignment is therefore the largest single sequential increment in this ablation, but the gains are distributed across several complementary terms rather than being attributable to edge guidance alone. The no-edge interaction control reaches $\estdata{0.919\pm0.042}$, 0.009 below the full configuration, providing additional evidence that the MRI-derived cue contributes when reslice consistency is already present.

\subsection{Cross-Literature Positioning}
Table~\ref{tab:literature_context} provides numerical context from representative recent work. These values are copied from the corresponding publications and are \emph{not} directly comparable with the within-study tables: datasets, anatomy, input views, training populations, and metric definitions differ. GHD+DVS is additionally rerun within our study protocol, whereas the other literature values are used only for methodological positioning.

\begin{table*}[t]
\centering
\caption{Cross-literature numerical context. Values are publication-reported and are shown only for positioning; no cross-row ranking is claimed.}
\label{tab:literature_context}
\scriptsize
\begin{tabular}{p{2.2cm}p{2.4cm}p{3.0cm}p{3.7cm}p{3.6cm}}
\toprule
Method & Dataset / task & Input / supervision & Representative published result & Why not directly comparable \\
\midrule
GHD+DVS~\cite{luo2026ghdheart} & ACDC / UKBB sparse cardiac mesh fitting & sparse 2D segmentations; per-case template optimization & published summary reports approximately $0.90$ Dice for sparse fitting & different datasets, anatomy and mesh objective \\
Slice2Mesh~\cite{xiao2025slice2mesh} & clinical cine CMR; LV surface & sparse SAX/LAX images + partial contours; cohort training & mean Chamfer distance $3.621$ mm on 150 test samples & learned cohort model; CD rather than our 3D Dice/HD95 \\
HybridVNet~\cite{gaggion2025hybridvnet} & UK Biobank; ventricular meshes & multi-view CMR + mesh/cohort supervision & LV-Myo Dice $0.84$; LV-Endo HD $3.89$ mm; LV-Myo MCD $1.35$ mm & dense cohort supervision and different structures \\
NIHC~\cite{muffoletto2026nihc} & UK Biobank-scale cohorts; biventricular anatomy & sparse segmentations; 5000-mesh training cohort & LVM Dice $0.91\pm0.04$; CVD surface ED $2.51\pm0.33$ mm & point/mesh Dice and large learned anatomical prior \\
NISF++~\cite{stolt2026nisfpp} & UK Biobank 120-subject study & SAX/LAX CMR; cohort implicit field & average in-plane LV blood-pool Dice $0.88\pm0.18$ & in-plane segmentation / 3D+time representation \\
Swin+GAT~\cite{abhishek2026swingat} & MM-WHS CT/MRI; direct mesh & full 3D image + cohort learning & reported MRI Dice $0.83$, mean CD $1.8$ mm, 95th-percentile surface distance $<5$ mm & whole-heart/full-image task rather than sparse LV weak supervision \\
\textbf{\method{} (ours)} & MM-WHS MR20; LV endocardium & 4--64 automatic weak masks + CMR; per case & sparse-16: TransUNet $0.906\pm0.042$; nnU-Net $\estdata{0.928\pm0.025}$; Medical SAM3 $\estdata{0.928\pm0.040}$ 3D Dice & study-specific sparse protocol \\
\bottomrule
\end{tabular}
\end{table*}

The literature survey therefore supports a task-specific novelty claim rather than an unrestricted SOTA claim: to the best of our knowledge, prior work has not reported the same combination of MM-WHS MRI, automatically generated sparse axial weak masks from multiple upstream segmenters, and patient-specific LV SDF reconstruction without a cohort-level shape prior. Published numbers from other protocols should not be used to claim a universal MM-WHS ranking.

\subsection{Qualitative and Failure Analysis}
\begin{figure*}[t]
    \centering
    \includegraphics[width=0.88\textwidth]{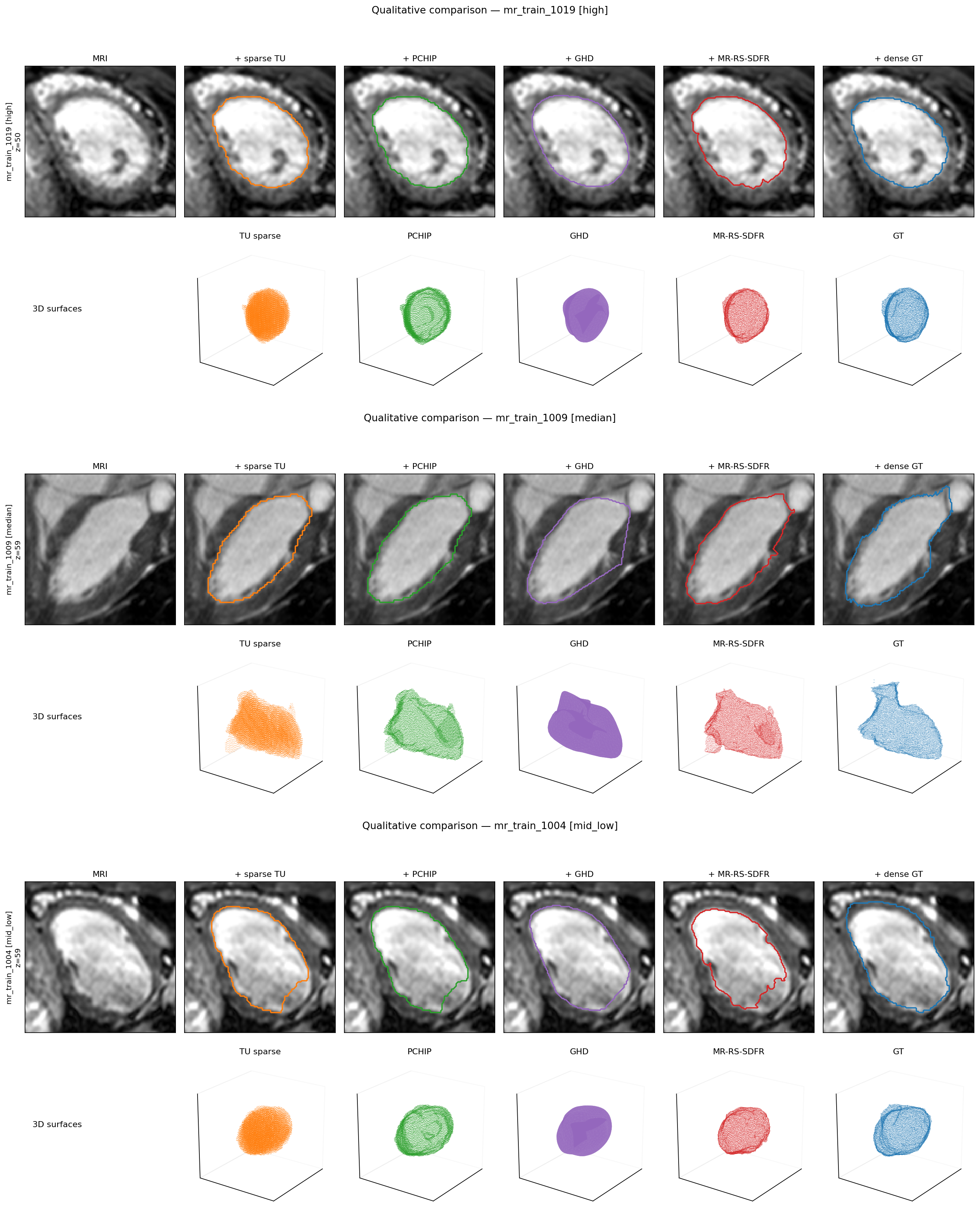}
    \caption{Performance-stratified qualitative comparison. The final figure should compare TransUNet-, nnU-Net-, and Medical-SAM3-derived sparse masks together with PCHIP, full GHD+DVS, MR-RS-SDFR, and dense reference contours/surfaces under matched views.}
    \label{fig:qualitative}
\end{figure*}

Figure~\ref{fig:qualitative} complements aggregate metrics with performance-stratified cases. The montage is used only to illustrate characteristic contour and surface behavior; quantitative claims are based on the 20-case aggregate results. The final version should render all three weak-mask sources and the corresponding sparse-16 MR-RS-SDFR outputs from final full-model checkpoints at $\phi=0$. Existing proxy panels may be retained only during manuscript development and must not be used as final-model evidence.

\subsection{Efficiency and Comparison Scope}
\begin{table}[t]
\centering
\caption{Runtime comparison on the same NVIDIA A800 80~GB GPU. Stage C is an evaluation/output step and is excluded from the reconstruction-time comparison.}
\label{tab:runtime}
\small
\begin{tabular}{lcc}
\toprule
Method / component & Time (s) & Reconstruction time \\
\midrule
MR-RS-SDFR Stage A & 66 & included \\
MR-RS-SDFR Stage B & 108 & included \\
MR-RS-SDFR Stage C & 18 & excluded (evaluation) \\
\midrule
MR-RS-SDFR (A+B) & \textbf{174} & included \\
Full GHD+DVS & 176 & included \\
\bottomrule
\end{tabular}
\end{table}

Table~\ref{tab:runtime} summarizes runtime on the same NVIDIA A800 80~GB GPU. MR-RS-SDFR requires approximately 66~s for Stage A and 108~s for Stage B, giving a reconstruction time of 174~s per case. Stage C requires an additional 18~s for evaluation/output generation but is excluded from the reconstruction-time comparison. Under the same hardware setting, full GHD+DVS requires 176~s per case. The two patient-specific reconstruction methods therefore have essentially the same computational cost in this implementation, while MR-RS-SDFR provides the accuracy advantages reported in Table~\ref{tab:sparse16_main}. Upstream weak-mask generation is kept separate from this timing comparison because TransUNet and nnU-Net require target-dataset LOO training, whereas Medical SAM3 is used from a pretrained checkpoint without MM-WHS-specific fine-tuning.

Within-study numerical ranking is restricted to methods evaluated under matched data and metric definitions. Dense segmentation rows are contextual references because they use full-volume predictions; PCHIP, Init, MR-RS-SDFR, and full GHD+DVS are sparse reconstruction comparisons when driven by the same selected masks. Cross-literature values remain positioning references only.

\section{Discussion}

The expanded results support five main conclusions. First, the proposed reconstruction framework is not tied to a single weak-mask generator. TransUNet, nnU-Net, and Medical SAM3 differ substantially in training regime and direct segmentation accuracy, yet each can provide sparse observations for the same patient-specific SDF optimization. This broadens the original formulation from a TransUNet-specific pipeline to a generator-agnostic reconstruction framework.

Second, upstream segmentation accuracy and downstream reconstruction utility are related but not identical. The direct 2D Dice ordering is nnU-Net $>$ Medical SAM3 $>$ TransUNet, whereas dense-volume 3D Dice/HD95 orders nnU-Net $>$ TransUNet $>$ Medical SAM3. After sparse-16 MR-RS-SDFR reconstruction, nnU-Net and Medical SAM3 both reach 0.928 mean Dice, with nearly identical HD95 of 3.81 and 3.80 mm, while TransUNet reaches 0.906 Dice and 6.41 mm. The aggregate results therefore support generator robustness and convergence toward a similar downstream accuracy for the two stronger sparse-mask sources, but they do not support a strict Medical-SAM3-versus-nnU-Net ranking without paired per-case analysis.

Third, the sparse-density study separates interpolation, geometric initialization, and MRI-guided refinement. PCHIP and Init remain sensitive to observation density, and under sparse-4 supervision Init is not consistently better than PCHIP. Final MR-RS-SDFR improves sharply up to sparse-16 and then exhibits essentially unchanged mean Dice through sparse-64 for all three generators. Surface-distance behavior is less monotonic beyond sparse-16, so the data support saturation rather than continuous improvement with additional planes. This is consistent with the intended role of the MRI cue: it supplies information beyond slice interpolation, while reslice consistency ties the field to the observed planes once a moderate amount of geometry is available.

Fourth, the Medical-SAM3 sparse-16 factorized analysis supports complementary roles for the Stage-B terms. Shared stabilization contributes +0.006 Dice, MRI edge-field alignment provides the largest single sequential increment (+0.008), reslice Dice adds +0.005, and contour consistency adds +0.002. The no-edge interaction control remains 0.009 below the full configuration. These results support a meaningful MRI-edge contribution without attributing the entire refinement gain to that term alone.

Fifth, full GHD+DVS provides a stronger explicit-mesh comparator than the earlier sparse-occupancy fit. MR-RS-SDFR is numerically better in both Dice and HD95 for TransUNet-, nnU-Net-, and Medical-SAM3-derived sparse-16 masks. The Dice margins are modest for TransUNet (+0.007) and especially nnU-Net (+0.002), but are larger for Medical SAM3 (+0.048); the corresponding HD95 reductions are 3.87, 4.88, and 5.43 mm. Because both reconstruction methods receive exactly the same sparse masks in each matched comparison, this isolates the downstream reconstruction formulation more directly than cross-dataset literature comparisons, while paired tests remain necessary for significance claims.

Medical SAM3 also changes the supervision narrative in a useful but specific way. The Medical SAM3 model itself is pretrained on external medical data, so it should not be described as an untrained model. In this study, however, it requires no MM-WHS-specific training, fine-tuning, or LOO model fitting.

The fixed-zero-level evaluation removes one potential source of hidden tuning: occupancy is defined by $\phi<0$ and the mesh is extracted at $\phi=0$, with dense labels used only after reconstruction. The within-study tables report HD95 in physical millimetres using a common evaluation definition. Final paired analyses should be regenerated from the same prediction-to-physical-space evaluation path to preserve this comparability.

Finally, the per-case formulation has a computational cost comparable to the closest patient-specific explicit-mesh baseline. As summarized in Table~\ref{tab:runtime}, Stage A and Stage B require 174~s in total, compared with 176~s for full GHD+DVS on the same A800 GPU. Stage C adds 18~s only for evaluation/output generation and is not included in this comparison. Thus, the observed accuracy gains over GHD+DVS are obtained without a material increase in patient-specific reconstruction time. Both approaches remain slower than feed-forward inference, while the Medical-SAM3 branch additionally avoids target-dataset training for weak-mask generation.

\section{Limitations and Future Work}
\label{sec:limitations}

The study has several limitations. First, evaluation remains restricted to 20 labeled MM-WHS MR training cases, so the results do not establish generalization across institutions, scanners, or external pathologies. Second, the three weak-mask generators are not identical in supervision regime: TransUNet and nnU-Net are trained with MM-WHS labels under LOO splits, whereas Medical SAM3 is pretrained externally and used without target-dataset fine-tuning. The comparison is therefore intentionally a weak-mask-source study rather than a claim of equal training supervision. Third, Medical SAM3 is prompt driven; its final prompt protocol must be fully specified and held fixed across cases for reproducibility. Fourth, nnU-Net and Medical SAM3 have the same sparse-16 mean final Dice at the reported precision, so their relative downstream ranking cannot be resolved from aggregate means alone and requires paired per-case analysis. Fifth, the protocol-matched GHD+DVS experiments require careful adaptation of the published template-based method to all three sparse-16 mask sources and should not be generalized into a universal ranking against every published GHD+DVS configuration. Sixth, the current qualitative figure does not yet show final full-model renders for all generators and representative failure cases. Finally, per-case optimization remains slower than feed-forward reconstruction, although its measured reconstruction time is comparable to full GHD+DVS under the same hardware setting.

Future work will emphasize external multi-center validation, expert sparse contours, additional foundation-model mask generators, final-checkpoint failure-case visualization, and acceleration of Stage-B optimization beyond the current GHD+DVS-comparable runtime. Clinical extensions include LV volume error and Bland--Altman analysis, while methodological extensions include RV/myocardial and whole-heart reconstruction. For multi-structure SDF reconstruction, S2MDF-style inter-object constraints~\cite{mercadier2026s2mdf} provide a natural mechanism for preventing anatomically implausible intersections.

\section{Conclusion}

We presented \method{}, a cross-slice implicit SDF framework for reconstructing the 3D LV endocardial surface from cardiac MRI under sparse axial weak-mask supervision. Strong geometric initialization establishes a continuous prior, MRI edge-field normal alignment supplies an image-derived boundary cue independent of the weak masks, and differentiable reslice losses maintain consistency with the acquired planes. The expanded evaluation uses three automatic weak-mask generators---LOO TransUNet, LOO nnU-Net, and pretrained Medical SAM3 without MM-WHS-specific training or fine-tuning---and five sparse-view variants from 4 to 64 axial planes.

In the sparse-16 setting, MR-RS-SDFR reaches $0.906\pm0.042$ Dice and $6.41\pm4.70$ mm HD95 with TransUNet masks, $\estdata{0.928\pm0.025}$ and $\estdata{3.81\pm2.25}$ mm with nnU-Net masks, and $\estdata{0.928\pm0.040}$ and $\estdata{3.80\pm2.49}$ mm with Medical SAM3 masks. The direct generator results show different 2D and dense-3D orderings, while the final nnU-Net- and Medical-SAM3-driven reconstructions converge to the same mean Dice. Across all three sparse-16 mask sources, MR-RS-SDFR is numerically better than protocol-matched full GHD+DVS. The 4/8/16/32/64-plane progression shows a marked gain up to sparse-16 followed by saturation of mean final Dice, indicating limited additional benefit from denser sparse supervision under the current configuration.

Factorized analysis in the Medical-SAM3 sparse-16 setting shows complementary contributions from shared Stage-B stabilization, MRI edge-field alignment, reslice Dice, and contour consistency; MRI edge provides the largest single sequential Dice increment, while the no-edge control remains below the full model. Taken together, these findings support a generator-agnostic view of patient-specific MRI-guided SDF refinement: the framework can use sparse masks from either target-trained segmentation networks or an off-the-shelf medical foundation model while preserving its core geometric and image-guided reconstruction mechanism.

\section*{Compliance with Ethical Standards}
This study uses the public MM-WHS challenge dataset. Sparse reconstruction uses automatically generated masks from LOO TransUNet, LOO nnU-Net, or a pretrained Medical SAM3 model used without MM-WHS-specific fine-tuning. Dense LV labels are reserved for evaluation rather than Stage-A/Stage-B SDF optimization.

\section*{Acknowledgments}
This work was supported by \fundingtext.


\balance
\end{document}